\documentclass[10pt,twocolumn,letterpaper]{article}

\usepackage[pagenumbers]{cvpr} %

\usepackage{bbding}
\usepackage{caption}  %
\usepackage{subcaption}  %
\usepackage{booktabs}  %
\usepackage{tikz}
\usepackage{tabularx}
\usepackage{textcomp}
\usepackage{color}
\usepackage{cuted}
\usepackage{etoolbox}
\usepackage{graphicx}
\usepackage{makecell}
\usepackage[T1]{fontenc}
\usepackage{siunitx}
\usepackage{amsmath}

\newcommand{\mypar}[1]{\vspace{1mm}\noindent {\bf #1}~~}

\newcommand{\soundparams}{\phi}
\newcommand{\gsparams}{\theta}

\newcommand{\fielda}[0]{F_{\soundparams}}
\newcommand{\fieldg}[0]{F_{\gsparams}}

\newcommand{\pt}[0]{\mathbf x}
\newcommand{\view}[0]{\mathbf r}
\newcommand{\rgb}[0]{\mathbf c}
\newcommand{\depth}[0]{\mathbf d}
\newcommand{\video}[0]{\mathbf v}
\newcommand{\action}[0]{\mathbf a}
\newcommand{\sound}[0]{\mathbf s}

\definecolor{cvprblue}{rgb}{0.21,0.49,0.74}
\usepackage[pagebackref,breaklinks,colorlinks,allcolors=cvprblue]{hyperref}

\def\paperID{2769} %
\def\confName{CVPR}
\def\confYear{2025}

\title{Hearing Hands: Generating Sounds from Physical Interactions in 3D Scenes} %

\vspace{-2mm}
\author{
Yiming Dou\textsuperscript{1} \quad
Wonseok Oh\textsuperscript{1} \quad
Yuqing Luo\textsuperscript{1} \quad
Antonio Loquercio\textsuperscript{2} \quad
Andrew Owens\textsuperscript{1} \quad
\vspace{2mm} \\
\textsuperscript{1}University of Michigan
\quad \textsuperscript{2}University of Pennsylvania
\\
}

\begin{document}
\maketitle

\begin{strip}
\centering
    \centering
    \raggedright
    \vspace{-10mm}
    \includegraphics[width=\linewidth]{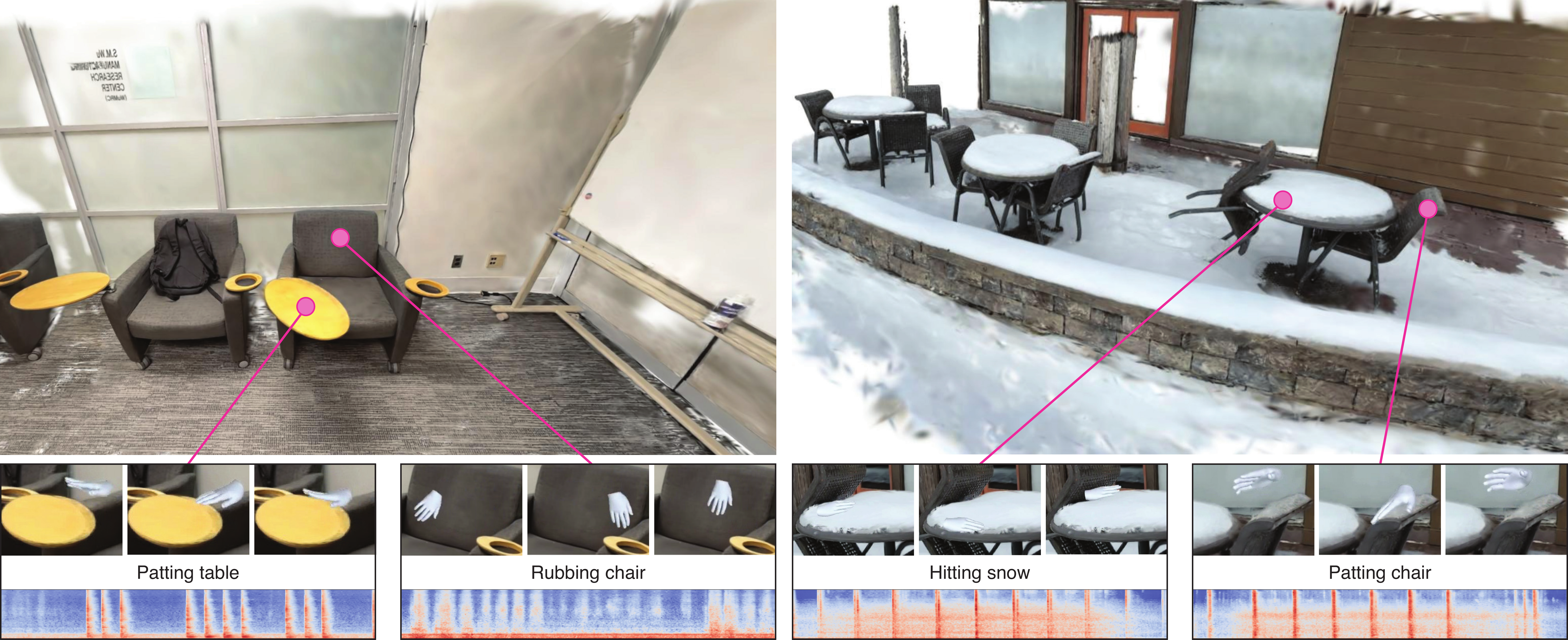}
    \captionof{figure}{
{\bf What sound does this object make when you strike it with your hand?} We capture a 3D scene representation that can be used to simulate the sound that would result from a given hand motion. We reconstruct the scene Gaussian Splatting~\cite{kerbl3Dgaussians}, then manipulate objects in the scene with hands, obtaining a sparse set of action-sound pairs. We use these examples to train a rectified flow model to map 3D hand trajectories at given position in a scene to a corresponding sound. At test time, a user can query an arbitrary 3D hand action and the model will estimate the resulting sound. Here we show several captured hand and audio pairs for two scenes (with representative video frames).}   
\label{fig:teaser}
\end{strip}

\begin{abstract}

We study the problem of making 3D scene reconstructions interactive by asking the following question: can we predict the sounds of human hands physically interacting with a scene?
First, we record a video of a human manipulating objects within a 3D scene using their hands. %
We then use these action-sound pairs to train a rectified flow model to map
3D hand trajectories to their corresponding audio. 
At test time, a user can query the model for other actions, parameterized as sequences of hand poses, to estimate their corresponding sounds.
In our experiments, we find that our generated sounds accurately convey material properties and actions, and that they are often indistinguishable to human observers from real sounds. Project page:  \small{\url{https://www.yimingdou.com/hearing_hands/}}.

\end{abstract}

\section{Introduction}

Today's 3D reconstruction methods~\cite{mildenhall2021nerf,kerbl3Dgaussians} generally represent scenes as collections of static objects. While these representations are well-suited to many computer vision applications, they lack the ability to model {\em physical interactions}, such as what would happen if we struck an object with our hands. Modeling these interactions is a core challenge in a number of domains, ranging from AR/VR to robotics.

\looseness=-1 An emerging line of work aims to address this problem, particularly by modeling action-conditioned visual dynamics, resulting in reconstructions where one can open and close a microwave, operate scissors, or animate an object~\cite{davis2015image,torne2024reconciling,garfield2024,kerr2024rsrd,xie2024physgaussian, jiang2024vr-gs}. %
While these approaches have been effective, they primarily focus on the visual and structural changes that objects undergo, and may not always be applicable to all objects, such as those that do not articulate or deform.

We focus instead on an aspect of interaction for 3D reconstruction that is complementary to these approaches: predicting the sound that an action would make if it were performed in a scene. Beyond making scenes more immersive and the interaction more realistic, studying the sounds of actions could provide a more complete understanding of the scene, beyond what's accessible from only its visual appearance~\cite{owens2016visually,kac1966can}.
For instance, the sound we obtain from interacting with a surface can tell us whether it is hard or soft, smooth or rough, and hollow or dense.  %
In addition, by predicting sound, one can implicitly model highly dynamic effects, such as vibrations or deformations of objects~\cite{zhang2017shape,davis2015image,davis2014visual}.

We aim specifically to create 3D reconstructions that enable us to predict what sounds a human hand will make when it interacts with the scene. We choose to parameterize our actions using hands, rather than alternatives such as drumstick~\cite{owens2016visually} or hammer strikes~\cite{gao2023objectfolder}, since they can execute a highly diverse range of actions by hitting, scratching, and manipulating objects. Hand sounds are also crucial for simulating interactions that a human might make in a virtual world application~\cite{nordahl2011sound}. Finally, the actions that a hand makes can be parameterized using trajectories of 3D hand reconstructions, which can easily be captured using ordinary video cameras~\cite{hasson2019learning,shan2020understanding,pavlakos2024reconstructing}.

We take advantage of the link between a material's visual appearance and the sound that it generates when it is physically manipulated~\cite{owens2016visually,zhang2017shape,du2023conditional}.
In contrast to vision-to-sound work, however, we are interested in generating the sound of user-specified \emph{simulated} interactions, without the need for an input video (Fig.~\ref{fig:teaser}). %
To do this, we collect a dataset of 3D hand-scene interactions paired with sounds. %
We first record a video where a person interacts with objects using their hands. We then estimate hand pose and register it to the same space as a 3D scene reconstruction, obtained using Gaussian Splatting~\cite{kerbl3Dgaussians} (Fig.~\ref{fig:teaser}).
This allows us to remove body occlusions from the training data (Fig.~\ref{fig:dataset_examples}) and to obtain 3D-consistent data augmentation by generating different views of the same interaction.
We use this data to train a model based on rectified flow~\cite{wang2024frieren,liu2022flow} that, from a sequence of 3D hand poses and visual content from the scene, can generate the sound resulting from the hand's motion (Fig.~\ref{fig:teaser}).

To help study this problem, we collect a dataset containing 24 indoor and outdoor 3D scenes and 9.1 hours of physical interactions. Through our experiments on this dataset, we find that our model successfully generates sounds that convey hand motion, such as by capturing the timing of contact. These experiments also suggest that the generated sounds convey material properties of objects in the scene.

\section{Related Work}
\mypar{Multimodal 3D scene reconstruction.}
 A variety of recent works augment 3D reconstructions with other modalities. LERF~\cite{kerr2023lerf} distills CLIP~\cite{radford2021learning} features into a NeRF~\cite{mildenhall2021nerf}, which can be used in downstream tasks such as 3D visual grounding~\cite{yang2024llm} and task-oriented grasping~\cite{rashid2023language}. ObjectFolder~\cite{gao2021objectfolder,gao2022objectfolder,gao2023objectfolder} constructs multimodal representations for objects. However, they only consider small object-level reconstructions of rigid objects that can be captured with a special apparatus (e.g., a turntable) and are limited to simple impact sound. In contrast, our goal is to produce scene-level reconstructions and to support complex actions represented by hand motions. Tactile-augmented radiance fields~\cite{dou2024tactile} register sparse tactile signals into the 3D space, allowing one to query how a given 3D location would feel if touched. We consider sound instead of touch, and crucially we do not treat sound as an intrinsic property of a surface (like they do with touch). Instead, it is a function of the action that is applied to the scene, which is specified via a 3D hand trajectory.

\mypar{Material properties in 3D scene reconstruction.} Another line of works focuses on integrating dynamics into 3D scene representations. Early work~\cite{davis2015image} used modal models to simulate deformation. D-NeRF~\cite{pumarola2021d} augments a NeRF with a displacement field, which adds temporal information to the NeRF. Recently, PhysGaussian~\cite{xie2024physgaussian} uses explicit 3D Gaussian Splatting~\cite{kerbl3Dgaussians} to model the dynamic behaviors, and VR-GS~\cite{jiang2024vr-gs} further develops a dynamics-aware interactive Gaussian Splatting representation. 
Like these works, we model how a scene will react to a physical interaction. However, we focus on hand-based actions  and predict sound rather than visual deformation. Sound prediction provides a complementary way to analyze physical properties, especially in cases where visual deformation is not available (such as for hard surfaces).

\mypar{Video-to-audio generation.}
There have been many approaches for synthesizing audio from visual or language inputs. Early work predicted simple speech from vision~\cite{ngiam2011multimodal}. Our approach is closely related to work that generates sound as a way to understand material properties~\cite{owens2016visually,zhang2017shape,du2023conditional}. Early work in this area predicted sound from videos of a drumstick striking objects~\cite{owens2016visually}. In contrast, our input is a 3D trajectory of a hand, allowing us to query the model with user-specified actions at test time (without need for a video input), we trained with many samples within a single scene, and we use 3D constraints, such as to obtain a clear view of the action and materials. Later work used more powerful generative models for conditional audio generation, such as autoregressive models~\cite{zhou2018visual}, GANs~\cite{chen2020generating}, and VQ-GANs~\cite{iashin2021taming}. %
Recent work uses diffusion models. Diff-Foley~\cite{luo2024diff} represents the video using a joint audio-visual embedding~\cite{arandjelovic2017look,owens2018audio} from the video and  generates a sound using latent diffusion. Frieren~\cite{wang2024frieren} uses rectified flow matching~\cite{liu2022flow} for better generation quality and efficiency.
Our audio generation module is based on the Frieren's rectified flow matching, but we use conditional information from a sequence of 3D hand poses and visual content extracted from a Gaussian splatting representation, instead of predicting sound from a video.  

\mypar{3D audio reconstruction.} 
A recent line of work has generated sound from 3D body pose~\cite{xu2024sounding,huan2024modeling}. In contrast, we model the combination of the action and the real-world objects that it is physically interacting with, rather than the body itself. Work on acoustic reconstruction~\cite{su2022inras,du2021learning,majumder2022few,liang2023av,chen2023novel,chen2024real} models how a sound propagates through a 3D scene, given the position of a sound and a listener. This line of work is complementary to ours: we model the generated sound in a scene, rather than the interaction between the listener and the sound.

\section{Method}
\begin{figure}[t]
    \centering
    \includegraphics[width=\linewidth]{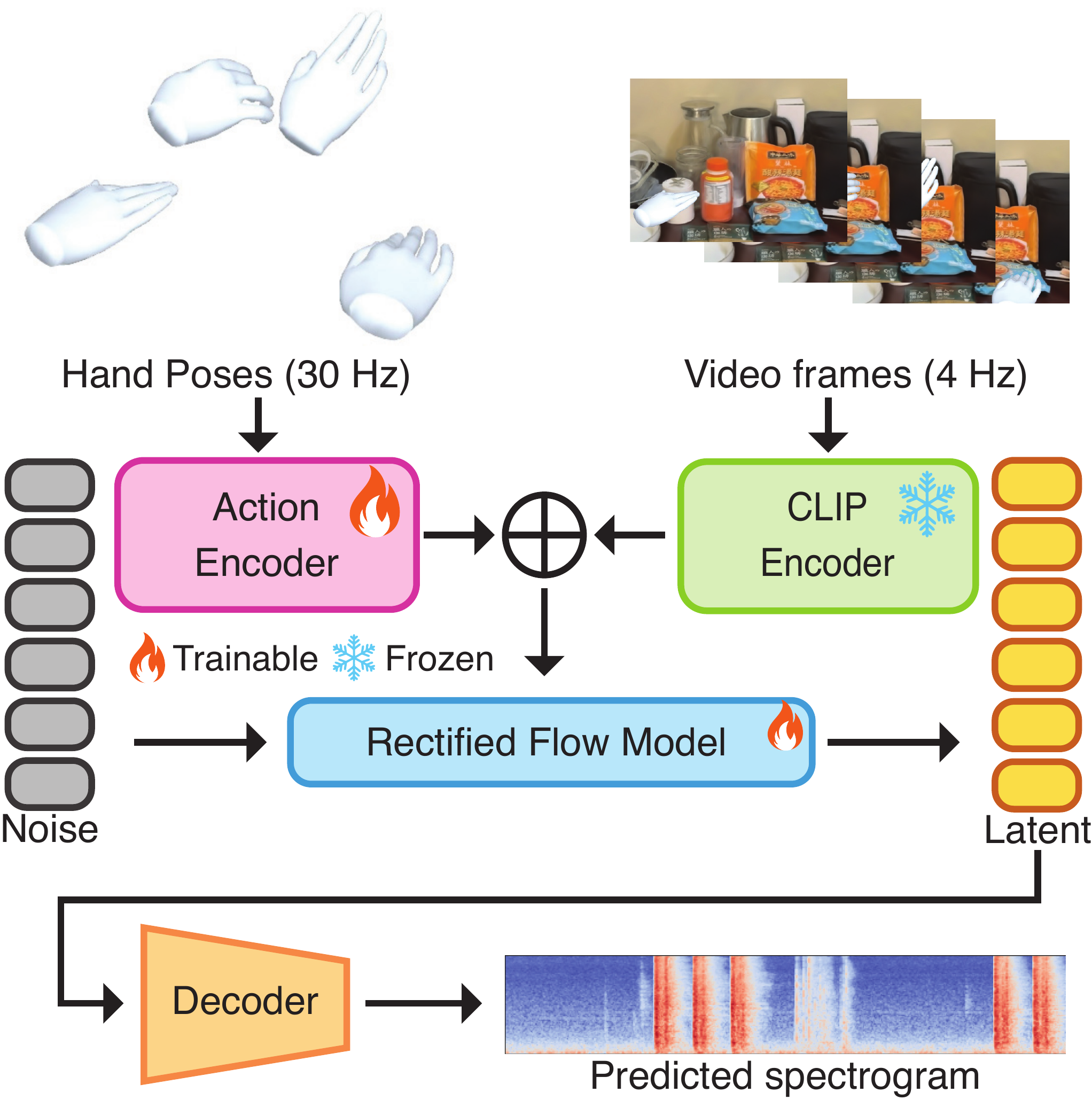}
    \caption{{\bf Sound generation}. We train a rectified flow model~\cite{wang2024frieren} to generate a sound spectrogram from a sequence of 3D hand positions and video frames generated from a 3D reconstruction of a scene. The sound can subsequently be converted into a waveform using a vocoder.}\vspace{-3mm}
    \label{fig:method}
\end{figure}

We aim to obtain a multimodal 3D reconstruction of a scene that allows us to predict the sound of actions.
To do so, we combine a visual neural field $\fieldg: (\pt, \view) \mapsto (\rgb, \depth)$ that maps a 3D point $\pt$ and viewing direction $\view$ to its corresponding RGB color $\rgb$ and depth $\depth$ with an action-conditioned audio estimator $\fielda: (\video, \action) \mapsto \sound$, which generates sound $\sound$ given the video $\video$ and the action $\action$. This action specifies the trajectory of a hand that physically interacts with the scene. We focus on human hands since they are capable of many motions (e.g., tapping, scratching, patting); they are crucial within virtual world applications; and can be easily captured in 3D without special equipment.

In the rest of this section, we explain how to generate a large and diverse dataset to train $\fielda$ (Sec.\ref{sec:dataset}). Then, we explain the functional form that we use to instantiate $\fielda$ (Sec.~\ref{sec:generative_model}).

\begin{figure*}[t]
    \centering
    \includegraphics[width=\linewidth]{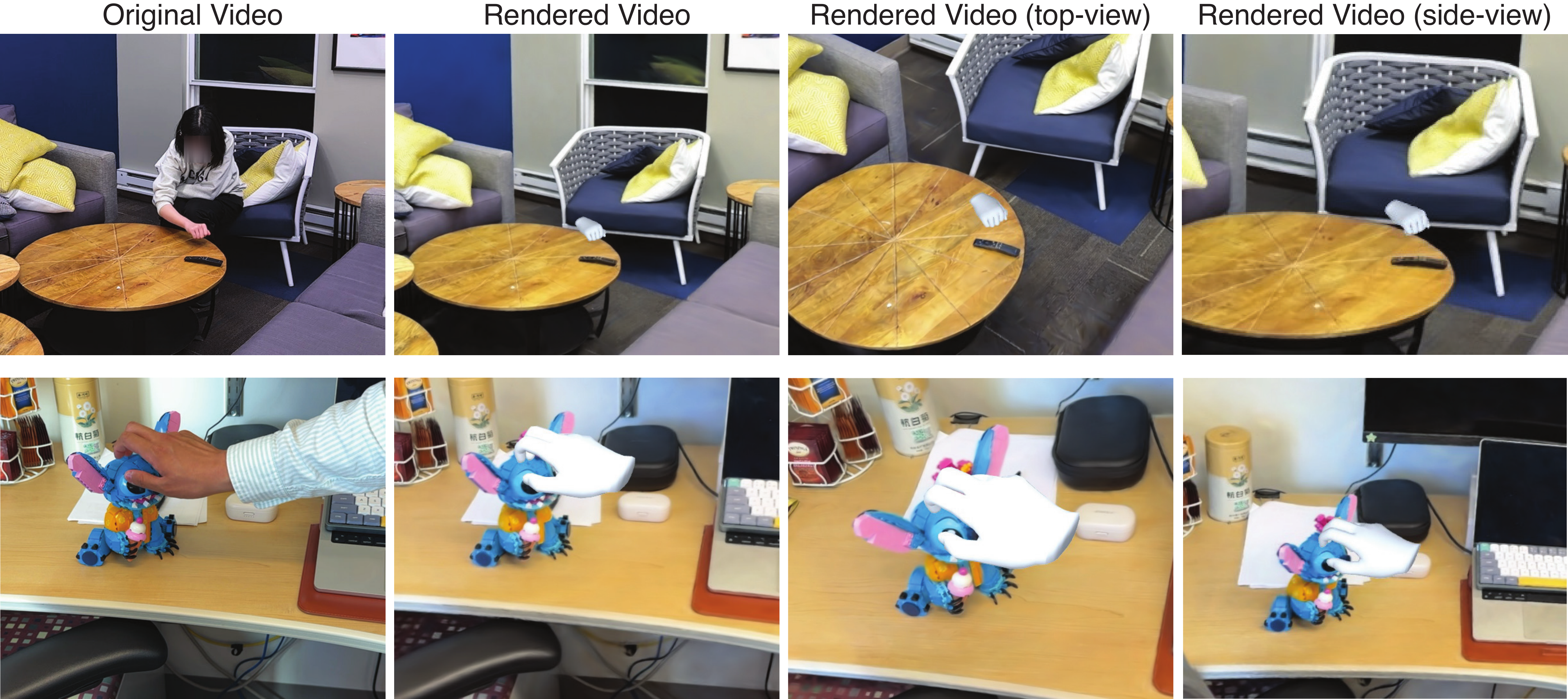}
    \caption{\textbf{Data capturing pipeline.} In the original video, a human collector interacts with the scene by performing various actions with their hands. We lift the annotator's hands to the same 3D space of the scene reconstruction. We render a video of the interaction by projecting 3D hands on multiple viewpoints of the scene. All rendered videos are synchronized with the sounds made by the hand actions.}
    \label{fig:capturing_pipeline}
\end{figure*}

\subsection{Dataset}
\label{sec:dataset}

\begin{figure*}[h]
    \centering
    \includegraphics[width=\linewidth]{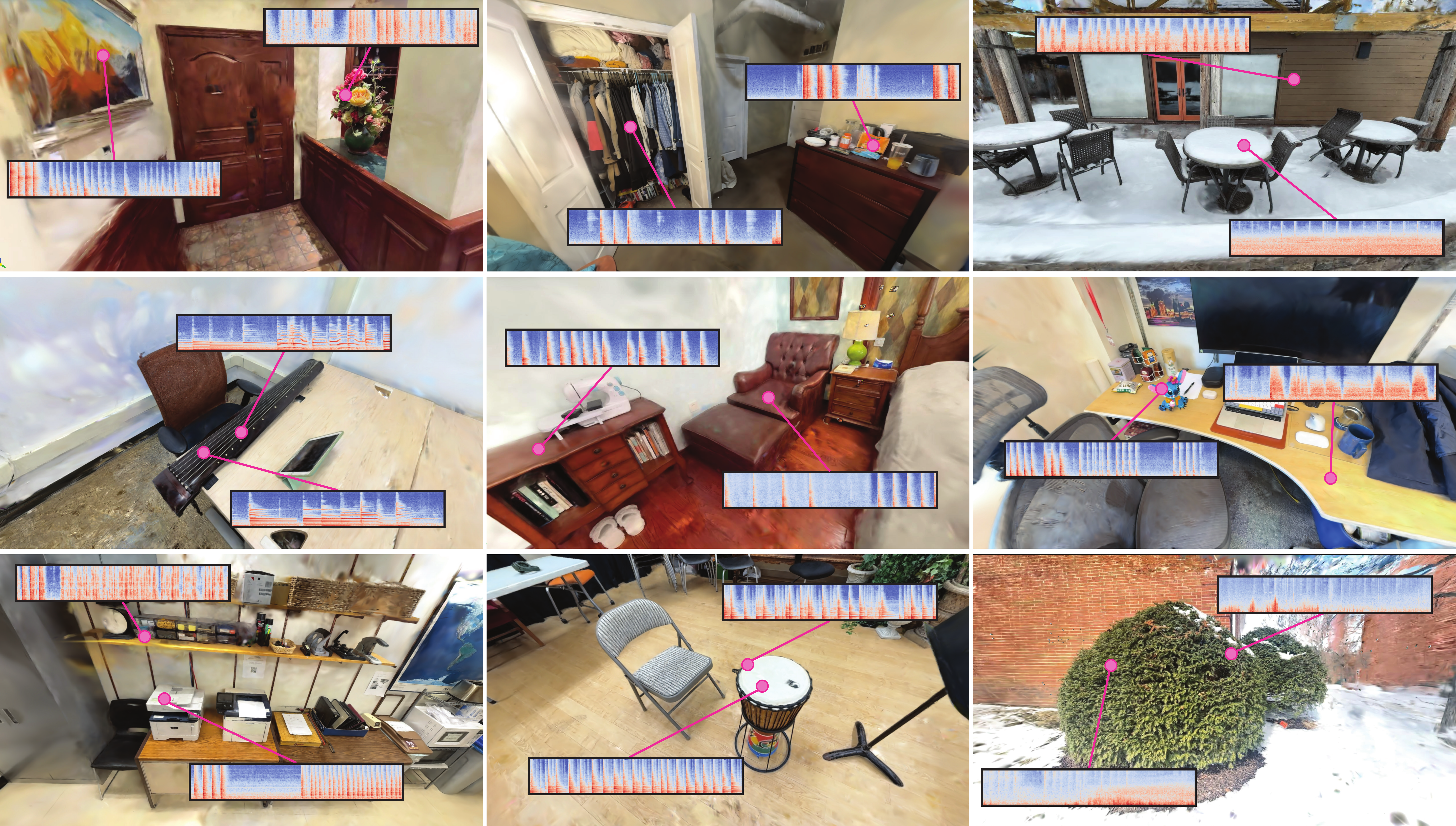}
    \caption{\textbf{Representative examples from the dataset.}
    Our dataset is collected in 24 scenes, including offices, outdoor trees, bedrooms, \etc. We show six such scenes in the figure above, with examples of action-generated sounds. Our dataset covers a wide range of actions (hitting, scratching, patting, \etc) and interacted materials (wood, metal, plastic. \etc). In each scene, approximately 1,400 seconds of videos are collected, resulting in a total of 9.1 hours of interaction data.}
    \label{fig:dataset_examples}
\end{figure*}

Training a generalizable $\fielda$ requires a diverse dataset of synchronized interaction videos $\video$, actions $\action$, and resulting sound $\sound$.
We collect this dataset in 24 different scenes, including bedrooms, lobbies, trees, snow, and musical instruments (see Fig.~\ref{fig:dataset_examples} for some dataset samples).
For each scene, we first generate a 3D reconstruction $\fieldg$ using Gaussian Splatting~\cite{kerbl3Dgaussians}. Specifically, a human collector scans the scene by recording multiple views, whose poses are estimated using the structure of motion~\cite{schoenberger2016sfm}.

After scanning, we collect videos of humans interacting with different regions of the scene (Fig.~\ref{fig:capturing_pipeline}).
During such interactions, the data collector performs a variety of actions with their hands, \eg, squeezing, hitting, or scratching, on some of the objects present in the scene, \eg, tables, plastic bags, or trees.
We use this procedure to generate a set of videos with various impact sounds.
Note that during each interaction, we keep the camera location fixed by mounting the recording device to a tripod.

We use HaMeR~\cite{pavlakos2024reconstructing} for 3D hand detection in such interaction videos. Specifically, we define the sequence of $N$ 3D hand keypoints for both hands as $\action \in \mathbb {R}^{2N\times21\times3}$. If one hand is not visible, we pad its detections with zeros. We register the camera on the tripod $c$ to $\fieldg$ with COLMAP~\cite{schoenberger2016sfm}, obtaining its global position $T_c^{\fieldg}$. 
Then, we use $\action$ and $\fieldg$ to generate a simulated interaction video $\video$. 
Specifically, we project the sequence of 3D hands $\action$ on an global RGB view of $\fieldg$ at the camera position $T_c^{\fieldg}$ (Fig.~\ref{fig:capturing_pipeline}).
We also re-center the camera position to each hand in $\action$ to obtain a sequence of local RGB views, which contains the local details of the regions being interacted with.
The simulated video $\video$ represents the combination of both global RGB views $\video_g$ with hands and local RGB views $\video_l$.
We label each $\video$ with the sound $\sound$ from the original video of the human interacting with the scene.

We collect approximately 1,400 seconds of videos in each scene, with a frame rate of 30Hz. We pre-process these videos to generate $\action$, $\video$, and $\sound$ as explained above. This pre-processing results in a dataset of approximately 9.1 hours of simulated interactions.
We additionally use the relative position of the camera to the scene $T_c^{\fieldg}$ to project $\action$ from the local camera frame to the global frame of $\fieldg$. This allows us to synthesize two novel views of the simulated interactions from slightly different viewpoints, \ie, top view, side view. Fig.~\ref{fig:capturing_pipeline} shows some representative samples for this process. To the best of our knowledge, this is the first dataset to capture human actions along with their sounds that are spatially aligned in 3D scenes.

\subsection{Generating action-conditioned sound}
\label{sec:generative_model}

We represent $\fielda$ as a generative model $p_{\soundparams}(\sound\mid~\video, \action)$ where $\sound$ is the sound generated by $\action$ in the video $\video$. Similarly to previous work, we represent $\sound$ as a mel-spectrogram, transforming audio synthesis into image generation.

We instantiate $p_{\soundparams}(\sound\mid~\video, \action)$ as a rectified-flow matching generative model~\cite{liu2022flow}.
Our model is built upon the video-to-sound Frieren model~\cite{wang2024frieren}. Similarly to Frieren, we compress $\sound$ to a latent vector with a pre-trained autoencoder, and train a generative model in latent space. However, we empirically found the Frieren model to fail to generate high-quality sound from our videos, even when finetuned on our dataset. This is because our videos contain simulated interactions, which lack the low-level details and consistency of real videos, e.g., the motion and deformation of objects.
Therefore, we introduce two key modifications to Frieren: (i) we encode $\video$ with CLIP~\cite{radford2021learning} instead of CAVP~\cite{luo2024diff} since we found CLIP to have better spatial consistency and material understanding; and (ii) we explicitly condition the model on 3D action $\action$, which forces the model to focus on the low-level details of the hand motion.  We empirically found these two modifications to be crucial for performance, as we demonstrate in the experimental section. A visualization of the schematics of our model can be found in Fig.~\ref{fig:method}. Further implementation details can be found in Sec.~\ref{sec:implementation details}.

We train $\fielda$ from scratch on our dataset. After training, we can generate the sound of previously unseen interactions $\hat{\action}$ in the scene $\fieldg$ by first selecting a camera viewpoint $T_c^{\fieldg}$ and then rendering a video of the interaction $\hat{\video}$. We then use our model to predict the interaction's sound $\hat{\sound}$ by passing $\hat{\action}$ and $\hat{\video}$ to our generative model. We use the ability to generate sound for new actions in the scene to design an interactive interface for $\fieldg$ (Sec.~\ref{sec:results}).

\section{Implementation Details}
\label{sec:implementation details}

We reconstruct the 3D scene using the Splatfacto method from Nerfstudio~\cite{nerfstudio}. Approximately 1K images taken from various views are used for each scene. The gaussians are randomly initialized with scale regularization~\cite{xie2024physgaussian}. During training, we optimize the reconstruction with the Adam~\cite{kingma2014adam} optimizer for 20,000 steps on a single NVIDIA RTX 2080 Ti GPU.

\subsection{Audio generation model training and inference}
Our implementation of $\fielda$ is based on Frieren~\cite{wang2024frieren} but differs on the conditioning module to better suit our task.
First, we use CLIP features instead of CAVP features for encoding the simulated interaction video $\video$. Specifically, we pass the global video $\video_g$ and local video $\video_l$ separately into the CLIP model and obtain two features, which are then concatenated into the input feature of our model. Note that, similarly to Frieren, we condition the model on the frames from the video down-sampled at 4Hz.
We also find that the visual features extracted from downsampled videos are insufficient to capture fine-grained hand motions present in our data.
Therefore, we additionally condition the model on the action $\action$,  which includes the trajectory of 3D hand poses. Being sampled at 30Hz, $\action$ gives the model a higher resolution view of the action.
We encode $\action$ to the same dimension of the frame embeddings via a linear layer, and normalize it to a unit vector. Finally, we upsample the frames and actions embeddings to the same temporal frequency of the sound spectrogram, \ie, 31.25 Hz, using nearest neighbor upsampling. We then obtain the final conditioning vector by summing the two embeddings elementwise. This conditioning vector is concatenated to the input noise and passed to the vector field estimator to generate the latent spectrogram representation of the sound.

Following previous works~\cite{luo2024diff,wang2024frieren}, we divide our dataset into non-overlapping chunks of eight seconds duration. The video's audio is downsampled to 16kHz and transformed into mel-spectrograms with 80 bins and a hop size of 256.
We use $10\%$ of the collected videos as the test set, $10\%$ as validation, and the remaining as the training set. We use the knowledge of each video's camera pose $T_c^{\fieldg}$ to ensure that none of the camera views in the test set overlap with the ones in the training and validation set.

\begin{table*}[t]
    \centering
    \small
    \setlength
    \tabcolsep{4pt}
    \caption{{\bf Ablation study.} Since CLIP features and hand poses respectively provide material information and precise sound synchronization, removing either of them from conditioning will result in a significant drop in the overall performance. In particular, removing CLIP features and hand poses results in the greatest drop in the CLAP \textit{material} accuracy and \textit{action} accuracy, respectively. Excluding synthetic-view data augmentation affects the performance generally.}
    \begin{tabularx}{\textwidth}{X l c c c c c c c c c}
        \toprule
          Model variation  & STFT $\downarrow$ & Envelope $\downarrow$& FID $\downarrow$& IS $\uparrow$& CDPAM $(\times 10^{-4}) \downarrow$ & \multicolumn{3}{c}{CLAP-acc $(\%) \uparrow$} & Labeled \textit{real} $(\%) \uparrow$\\\cmidrule(l){7-9}
           &  &  & && & \textit{all} &\textit{action} &\textit{material} &\\
        \midrule
            RegNet & $0.62$ & $0.77$ & $63.84$ & $5.73$ & $3.38$ & $1.08$ & $42.55$ & $3.52$ & - \\
            Frieren & $0.74$ & $0.81$ & $56.66$ & $16.76$ & $3.71$ & $23.94$ & $41.73$ & $42.55$ & $43.79 \pm 2.64$ \\
        \midrule
            Ours & $\textbf{0.50}$ & $\textbf{0.66}$ & $59.02$ & $\textbf{17.82}$ & $\textbf{3.32}$ & $\textbf{28.09}$ & $\textbf{50.50}$ & $\textbf{45.62}$ & $ 47.18 \pm 2.66$ \\
            ~~~ - w/o CLIP & $0.68$ & $0.77$ & $\textbf{58.07}$ & $17.10$ & $3.86$ & $18.25$ & $43.90$ & $31.80$ & $ 41.24 \pm 2.62$ \\
            ~~~ - w/o hand pose & $0.69$ & $0.77$ & $58.92$ & $16.76$ & $3.77$ & $20.96$ & $38.21$ & $39.11$ & $ 43.50 \pm 2.64$ \\
            ~~~ - w/o synthetic-view & $0.62$ & $0.73$ & $58.99$ & $17.42$ & $3.66$ & $24.12$ & $47.61$ & $40.56$ & $ 43.22 \pm 2.64$ \\
        \bottomrule
    \end{tabularx}
    \label{tab:ablation}
\end{table*}

We then train the model for 40 epochs with a batch size of 128 using the Adam~\cite{kingma2014adam} optimizer. We initialize the learning rate to $10^{-5}$, do a warmup to $4 \times 10^{-4}$ over 1000 steps, and finally linearly decrease it to $3.4 \times 10^{-4}$ over 22 epochs. We train on a single NVIDIA L40s.

At inference time, the model performs 26 sampling steps with a 4.5 guidance scale. The generated latent is then decoded into a mel-spectrogram with a pre-trained decoder~\cite{wang2024frieren}. Finally, a pretrained vocoder~\cite{lee2022bigvgan} is used to transform the spectrogram into a waveform.

\section{Experiments}

We design our experiments to answer the following questions: (1) Can $\fielda$ generate synthetic sounds that are almost indistinguishable from real ones? (2) How important is conditioning on $\video$ and $\action$? (3) Do the predicted sounds convey physical properties of the scene, \eg, its material and their position relative to the camera? 
We answer these questions with qualitative and quantitative experiments.

\begin{figure*}[h]
    \centering
    \includegraphics[width=\linewidth]{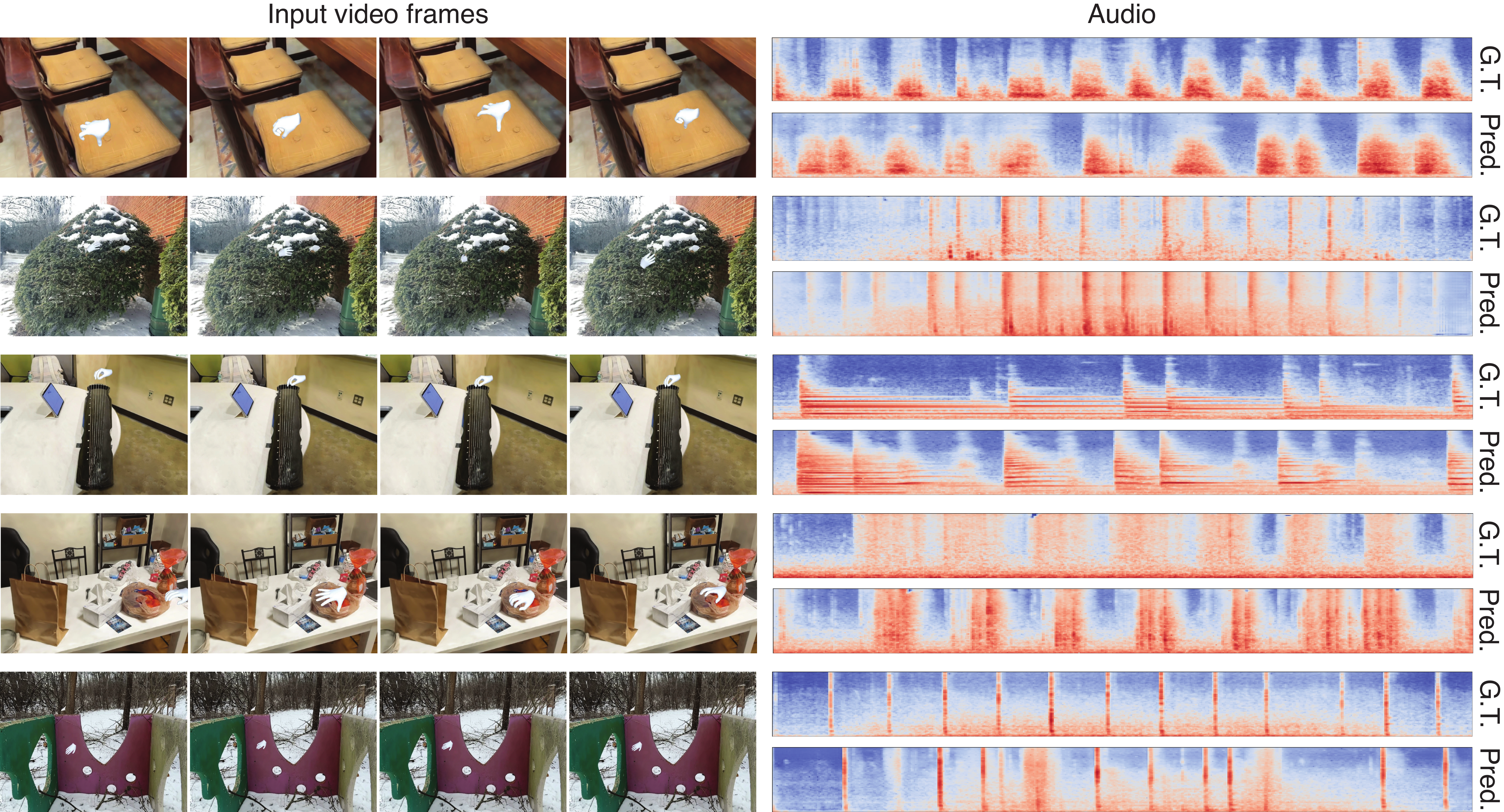}
    \caption{\textbf{Qualitative results.} We show the generation results on five interactions. Generally, the predictions match the ground-truth in both motion synchronization and material properties. Note that when the hand is less visible in the video or the motion is ambiguous (\eg, the last row), our model will generate less-synchronized audio spectrograms.
    }
    
    \label{fig:qualitative_results}
\end{figure*}

\begin{figure*}[h]
    \centering
    \includegraphics[width=\linewidth]{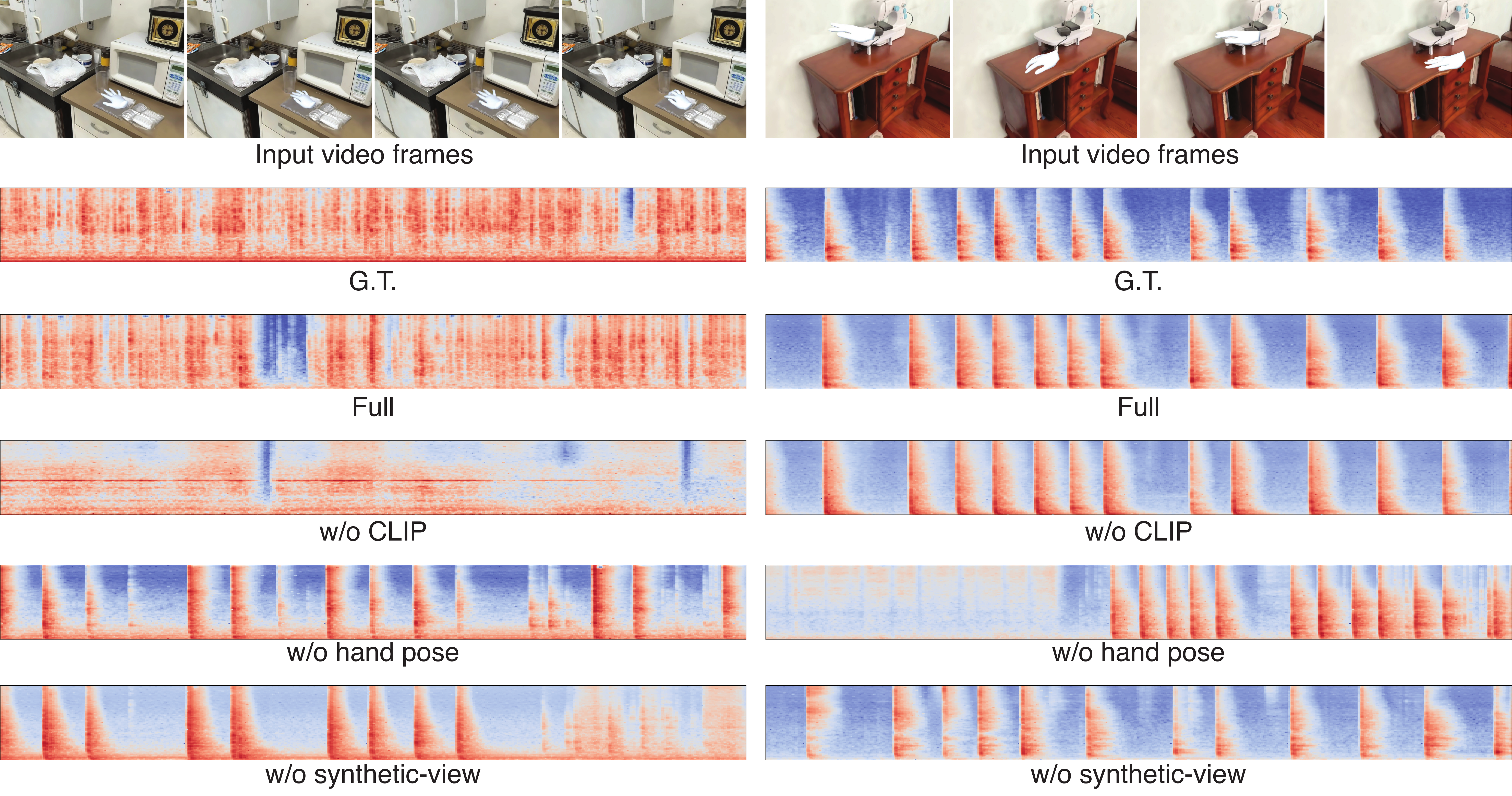}
    \caption{\textbf{Ablation study results.} We show the spectrogram predictions from our full model and three ablations. We notice that removing CLIP features softens impact sounds while removing hand pose features results in poor audio-video synchronization. Similar to quantitative results, the model trained without synthetic-view augmentation performs worse in both aspects.
    }
    
    \label{fig:ablation_qualitative_results}
\end{figure*}

\subsection{Experimental Setup}

We use the following metrics to evaluate the quality of the sounds generated by $\fielda$ and compare it to a set of baselines.

\mypar{Raw Audio Similarity.} 
As custom in previous work~\cite{gao2023objectfolder}, we measure the L2 distance between ground-truth and predicted audio signals in both the spectrogram (\emph{STFT}) and waveform (\emph{Envelope}) space. This metric primarily assesses the model’s capability to capture low-level sound features.

\mypar{Latent Space Similarity.} 
We encode both ground-truth and generated sounds to a latent representation and measure their distance in this space. Specifically, we adopt the CDPAM~\cite{cdpam} metric to measure distances in the latent space, which uses a pretrained model to quantify perceptual audio similarity. Additionally, following previous work~\cite{wang2024frieren}, we compute the Frechet Inception Distance (FID) and Inception Score (IS) using the pretrained mel-ception encoder model from SpecVQGAN~\cite{SpecVQGAN_Iashin_2021}.

\mypar{CLAP accuracy.}
To assess the model's effectiveness in generating sounds that accurately represent the actions and material properties in a scene, we introduce a new metric: \emph{CLAP accuracy}.
This metric evaluates whether an off-the-shelf CLAP model~\cite{laionclap2023} assigns the same zero-shot label to both the ground truth and synthetic sounds. Specifically, we define an action set $\mathbb{A}$ comprising 7 hand actions (e.g., knocking, scratching) and a material set $\mathbb{M}$ with 13 materials (e.g., wood, plastic). From these, we generate a set $\mathbb{P}$ of 91 action-material pairs by taking the Cartesian product of $\mathbb{A}$ and $\mathbb{M}$.
For each pair in $\mathbb{P}$, we format the CLAP model’s text prompt as: “This is a sound of hand \{action\} \{material\},” with \{action\} and \{material\} drawn from the pairs in $\mathbb{P}$. We then record the number of instances where the ground truth and generated sounds are assigned the same label (\emph{CLAP-acc, All}). For a more fine-grained analysis, we additionally report the frequency of action label matches (\emph{CLAP-acc, Action}) and material label matches (\emph{CLAP-acc, Material}).
This metric is inspired by prior work in sound generation~\cite{owens2016visually}, which similarly uses linear models to classify materials.

\mypar{Real-or-fake study.}
We conduct a real-or-fake user study to evaluate whether participants can distinguish between generated and real sounds. Fifty-nine participants participated in this study.
Each participant viewed 32 pairs of 8-second interaction videos $\video$ with each pair comprising one video with ground-truth sound and one with generated sound. These pairs were sampled from a set of 1107 video pairs, with sounds generated either by our full model or one of its ablations, selected with equal probability.
Following prior work~\cite{owens2016visually}, we use a two-alternative forced-choice (2AFC) test, where participants select the video they believe has the most realistic sound in each pair.
All videos in the study are from the test set.

\begin{figure*}[h]
    \centering
    \begin{subfigure}[t]{0.39\linewidth}
        \centering
        \includegraphics[height=3.4cm]{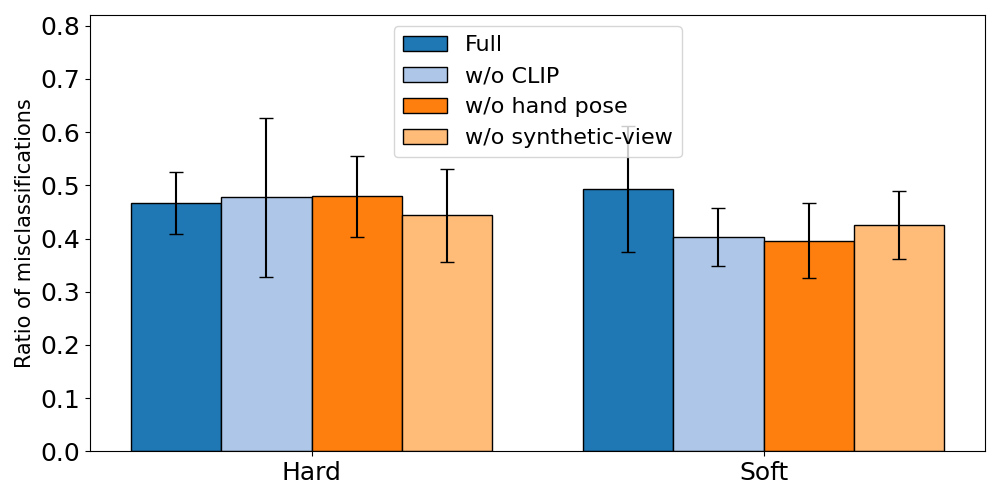}
        \label{fig:softness_acc}
    \end{subfigure}
    \hfill
    \begin{subfigure}[t]{0.39\linewidth}
        \centering
        \includegraphics[height=3.4cm]{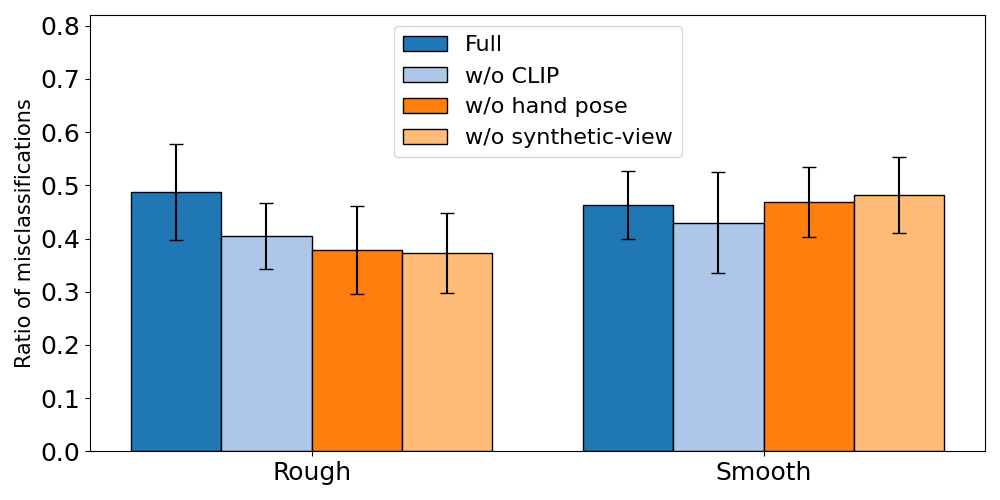}
        \label{fig:roughness_acc}
    \end{subfigure}
    \hfill
    \begin{subfigure}[t]{0.19\linewidth}
        \centering
        \includegraphics[height=3.4cm]{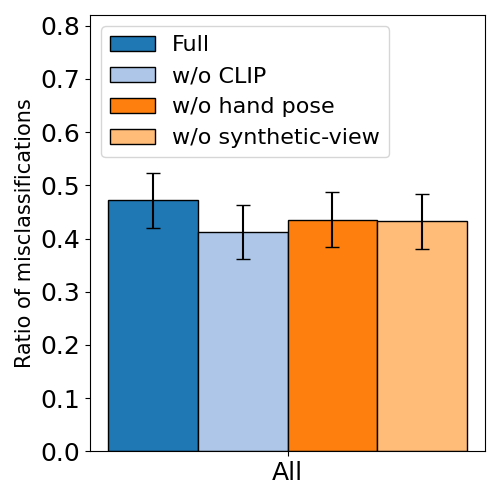}
        \label{fig:all_acc}
    \end{subfigure}
    \vspace{-5mm}
    \caption{\textbf{Results of real-or-fake study}. We show the ratio of humans being fooled by different variants of our model. We break down our results into three categories: softness, smoothness, and average over all samples. The error bars show $95\%$ confidence intervals. We find that our full model achieves a misclassification rate of approximately $47\%$, indicating the high quality of the generated sounds. In addition, our model generally outperforms baselines without visual or action information.}
    \label{fig:accuracy_comparison}
\end{figure*}

\subsection{Results}
\label{sec:results}

We begin by analyzing the differences in generated sounds produced by our full model and its ablations using quantitative distance metrics. The evaluation results, shown in Table~\ref{tab:ablation}, indicate that while all features of our model contribute to the generation quality, some are more essential than others.
Notably, removing conditioning on either the CLIP embeddings of the video or the 3D hand poses leads to a significant drop in performance. In contrast, excluding multi-view data augmentation during training has the smaller impact, resulting in relatively minor changes in both raw audio and latent distance metrics.
For metrics based on a pretrained melception model (FID and IS), all methods perform similarly. We hypothesize that this is due to our data differing significantly from VGGSound~\cite{chen2020vggsound}, the dataset on which the melception model was originally trained.

Interestingly, we observe that removing CLIP features results in the greatest drop in CLAP \textit{material} accuracy, while removing hand pose features most affects \textit{action} accuracy. This aligns with expectations: CLIP features primarily provide material information about the scene, while hand pose features are essential for encoding actions.

In Fig.~\ref{fig:qualitative_results}, we show qualitative results of our full model for five interactions. Visual inspection reveals that our model generates sounds that generally align with the ground-truth in both synchronization and material properties. We further show the qualitative results of ablation study in Fig.~\ref{fig:ablation_qualitative_results}.  We find that removing hand pose features disrupts audio-video synchronization, as visual information alone is insufficient for accurately estimating precise hand motions. Removing CLIP features, on the other hand, makes the model unable to synthesize the sound with correct material properties. Removing synthetic view results in general performance drop in both aspects.

\mypar{Real-or-fake study.}
We measure how often participants mistake our generated sounds for ground-truth sound.
We collect the data for this study on Amazon Mechanical Turk, obtaining answers from 59 participants.

Ideally, if the two sounds are completely indistinguishable from each other, we would observe a misclassification rate of $50\%$, which indicates that participants pick at random.\footnote{However, this is not an upper bound on performance, since subjects may sometimes prefer unrealistic sounds.}
The results of this analysis, averaged over all videos and participants, are shown in Tab.~\ref{tab:ablation}.
We find that our full system generates high-quality sounds with a misclassification rate of approximately $47\%$.

We present results broken down by the material properties of the objects the hand interacts with in Fig.~\ref{fig:accuracy_comparison}. Consistent with our quantitative findings, our approach outperforms all baselines on average. The improvement is especially notable for rough surfaces and soft materials, while differences are less pronounced for other categories.

Our study also suggested qualitatively interesting patterns in how users distinguish real from fake sounds. Notably, background noise in real recordings may sometimes be perceived as being artificial, whereas our model’s clearer outputs are often judged as more realistic. Users also sometimes may have been unfamiliar with the typical sounds of certain materials --- particularly those rarely encountered, such as snow -~- which can lead to inconsistent judgments. Additionally, inaccuracies in hand tracking and irrelevant movements during data collection can make it unclear whether the hand is interacting with the object or simply moving through space. This ambiguity might be mitigated by modeling object deformations resulting from contact.

\vspace{2mm}
\section{Conclusion}

We see our work as being a step toward creating realistic and immersive 3D scene reconstructions, with potential applications in robotics and AR/VR.
We do so by predicting the sound of hands interacting with a scene. Both automated evaluations and real-or-fake evaluations that our synthetic sounds outperform baselines and are often indistinguishable from real sounds. They also may convey material properties and subtle actions.

\mypar{Limitations.} One key limitation of our approach is that assumes that the objects in the scene do not move or deform when manipulated. In practice, this assumption is often violated, especially when manipulating small objects. 
  Another limitation comes from the errors of the 3D hand detection model, which might result in inaccurate hand motions in our dataset. This can be improved with future hand detection models.

\mypar{Acknowledgements.} 
We thank Jeongsoo Park, Ayush Shrivastava, Daniel Geng, Ziyang Chen, Zihao Wei, Zixuan Pan, Chao Feng, Xuanchen Lu, Ang Cao and the reviewers for the valuable discussion and feedback. 
We appreciate the generous help in data collection of Yueqi Ren, Qifan Wu, Shuangdi Zhang, Xingxian Li and Yuchen Huang from Qingyun Chinese Music Ensemble at UM.
We thank all the friends who participated in the real-or-fake study.
This work was supported by an NSF CAREER Award \#2339071, a Sony Research Award, and the DARPA TIAMAT program.

{
    \small
    \bibliographystyle{ieeenat_fullname}
    \bibliography{ref}
}

\end{document}